\documentclass[conference]{IEEEtran}
\IEEEoverridecommandlockouts
\usepackage{times}
\usepackage{epsfig}
\usepackage{graphicx}
\usepackage{amsmath}
\usepackage{amssymb}
\usepackage{subfig}
\usepackage{stfloats}
\usepackage{wrapfig}
\usepackage{cite}
\usepackage{amsmath,amssymb,amsfonts}
\usepackage{algorithmic}
\usepackage{graphicx}
\usepackage{textcomp}
\usepackage{xcolor}

\def\BibTeX{{\rm B\kern-.05em{\sc i\kern-.025em b}\kern-.08em
    T\kern-.1667em\lower.7ex\hbox{E}\kern-.125emX}}

\begin{document}

\title{FKIMNet: A Finger Dorsal Image Matching Network Comparing Component (Major, Minor and Nail) Matching with Holistic (Finger Dorsal) Matching}

\author{\IEEEauthorblockN{Daksh Thapar}
\IEEEauthorblockA{\textit{School of Computing and} \\
\textit{Electrical Engineering}\\
\textit{Indian Institute of Technology Mandi}\\
Mandi, India \\
d18033@students.iitmandi.ac.in}
\and
\IEEEauthorblockN{Gaurav Jaswal}
\IEEEauthorblockA{\textit{School of Computing and} \\
\textit{Electrical Engineering}\\
\textit{Indian Institute of Technology Mandi}\\
Mandi, India  \\
gaurav\_jaswal@projects.iitmandi.ac.in}
\and
\IEEEauthorblockN{Aditya Nigam}
\IEEEauthorblockA{\textit{School of Computing and} \\
\textit{Electrical Engineering}\\
\textit{Indian Institute of Technology Mandi}\\
Mandi, India \\
aditya@iitmandi.ac.in}
}


\maketitle

\begin{abstract}
Current finger knuckle image recognition systems, often require users to place fingers' major or minor joints flatly towards the capturing sensor. To extend these systems for user non-intrusive application scenarios, such as consumer electronics, forensic, defence etc, we suggest matching the full dorsal fingers, rather than the major/ minor region of interest (ROI) alone. In particular, this paper makes a comprehensive study on the comparisons between full finger and fusion of finger ROI's for finger knuckle image recognition. These experiments suggest that using full-finger, provides a more elegant solution. Addressing the finger matching problem, we propose a CNN (convolutional neural network) which creates a $128$-D feature embedding of an image. It is trained via. triplet loss function, which enforces the $L2$ distance between the embeddings of the same subject to be approaching zero, whereas the distance between any $2$ embeddings of different subjects to be at least a margin. For precise training of the network, we use dynamic adaptive margin, data augmentation, and hard negative mining.  In distinguished experiments, the individual performance of finger, as well as weighted sum score level fusion of major knuckle, minor knuckle, and nail modalities have been computed, justifying our assumption to consider full finger as biometrics instead of its counterparts.  The proposed method is evaluated using two publicly available finger knuckle image datasets i.e., PolyU FKP dataset and PolyU Contactless FKI Datasets.
\end{abstract}


\section{Introduction}
Biometric authentication, undoubtedly, has become the buzzword in the information security industry. Glancing at the potential impact it has on numerous privacy and protection applications as well as in our everyday life it can certainly be said that this innovative technology has the potential to revolutionize banking transactions, e-commerce, e-health and so on. Many features that make biometrics attractive have not just challenged the existing security systems, but have also revealed new security issues. 
Deployment of a biometric system based on hand dorsal characteristics may play a substantial role in instituting human identity in most of the real-time applications. The possible hand dorsal biometric traits are hand geometry or shape, finger geometry, nail bed, dorsal hand vein, and finger dorsal knuckles \cite{1}. Among them, the geometrical or shape features of hand/ finger are not very distinct to recognize individuals. While, the vein traits in hand are distinctive and difficult to spoof, but require extra hardware for mounting, lighting, and imaging devices \cite{5, 6}. On the contrary, the convex shape lines and skin folds on finger dorsal surface are very distinctive \cite{22}. Also, the acquisition procedures require relatively little user-cooperation and can be easily done using conventional low-resolution imaging cameras \cite{62}. Moreover, the finger dorsal knuckles are difficult to be rubbed and less prone to injuries unlike fingerprints because they exist on the outer side of the finger and are naturally preserved. Compared with hand-crafted design pipeline of traditional biometric approaches, deep learning, particularly the convolution neural network based approaches, automatically extract discriminative features from raw samples, and match the corresponding regions between two images. 

\begin{figure*}[!hbtp]
\small
\centering
\subfloat[Annotate Finger (Polyu FKI \cite{56})]{\includegraphics[scale = 0.4]{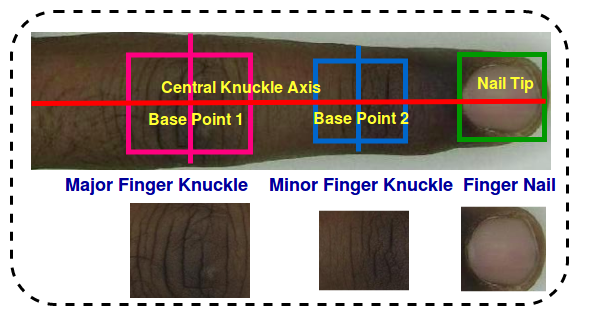}} %
\hspace{1 em}
\subfloat[Annotate Finger (PolyU FKP\cite{55})]{\includegraphics[scale = 0.28]{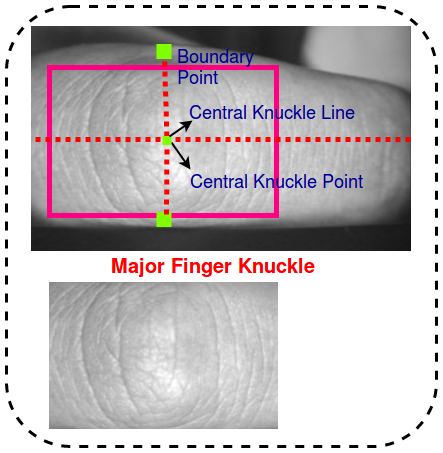}} %
\caption{Finger Knuckle Image Annotation}
\label{fig:1_1}
\end{figure*}

\subsection{Finger Knuckle: Concerns and Challenges}
The epidermal cells near the finger knuckle mature during an early stage of human development and seldom change during an adult’s life. These dorsal patterns are invariant to emotions or behavioural aspects and cannot be easily manipulated~\cite{1, 58}. However, there are still some tough challenges in the uncontrolled finger knuckle image (FKI) recognition which are yet to be solved. In existing works, researchers often combined major and minor finger knuckles for accurate matching but lack precise recognition results. Fig. \ref{fig:1_1} is depicting the annotated finger knuckle patterns on two different types of finger dorsal images. The componentwise finger knuckle ROI extraction is not consistent and tends to fail on some subjects because the finger features deviate during contact-less acquisition. One can't register with inconsistent ROI's until we exactly detect the middle knuckle line. Therefore, the open issues for practical FKI recognition to achieve real-time high performance are: (1) The large variation of finger dorsal changes require image detectors to accurately address a complicated finger and non-finger classification problem, (2) Majority of the former works didn't address the fingernail which has very unique shape (3) Varying illumination and background in outdoor conditions affect the matching. (4) Non-rigid distortion (5) Lack of non-uniform databases in which images incorporate the real world situations such as non-stretch palm or bending of fingers.

\textbf{Finger nail as biometric}: The part of the dermal structure underneath the nail plate remains robust and relatively invariant over the time (i.e, inner “U” shaped nail surface attach to the skin) \cite{65}. Whereas, the outer nailbed region can be altered easily by nail art, extensions or by nail cutting. Similar to any other convolutional network the FKIMNet is expected to address such issues by learning to ignore the variable region (outer nail) and emphasize more on the invariant region (inner nail) while authentication, with respect to the training dataset.

\subsection{Contribution}
We have detected three ROI components i.e., major knuckle, minor knuckle and nail in a finger dorsal region using state-of-the-art object detection technique and performed information fusion at score level for performance improvement. Each of the above finger component is very difficult to register consistently for an individual. This begs the following question: How feasible is a holistic finger dorsal image to identify individuals? To affirmatively address the above question, we present the contribution of our work in three-folds : (1) A novel full finger image-based matching network (FKIMNet) has been suggested and compared to the individual performance of major knuckle, minor knuckle and nail. In early works, no one has used finger holistically because the bigger finger size is difficult to handle. (2) To the best of our knowledge, the nail has been less used in the finger dorsal studies and also never fused with finger knuckles. However, nail has been found to be the best-performing finger dorsal component in contrast to the major and minor finger knuckles. (3) We propose FKIMNet architecture which takes highly rectangular augmented images as input and learns domain-specific line based features and their spatial relationship using tall and fat filters. In order to train the FKIMNet robustly, we select hard negative triplets during training using adaptive margin criterion for better discrimination, which mines hard triplets by increasing margin conditionally. For rigorous experimentation, we have involved two publicly available finger knuckle databases i.e., PolyU contact-less finger knuckle image (FKI) database \cite{56} and PolyU finger knuckle print (FKP) data-set \cite{55}. We are committed to release our trained network ($FKIMNet$) along with its training and testing source code modules which will be publicly available on $Github$ very soon.
 
The remainder of the article is organized into the following main sections: Section $2$ describes the related work in FKI recognition. The proposed biometric system which includes data augmentation, triplet loss, hard negative mining etc is elaborated in the $3^{rd}$ section. In section $4$, the experimental results are presented. Next, we discussed the various aspects of security analysis. Finally, the conclusion is drawn in the last section.

\section{Related Work}
In this section, the state of art studies related to ROI extraction, feature extraction and classification in the areas of finger knuckle is presented. However, the finger knuckle biometric is still a less investigated trait for a wide range of applications and a very few databases are available in the public domain for research and practices. The prime research in this area began in early 2009. 
\subsection{ROI Extraction}
Most of the state-of-the-art ROI segmentation algorithms considered convexity magnitude to locate the centre of the major finger joint \cite{26, 39}, when tested on publicly available PolyU FKP database. In another work, authors \cite{22,58} proposed ROI segmentation framework for utilizing information of Distal Inter Phalangeal (DIP), Proximal Inter Phalangeal (PIP) and Metacarpophalangeal (MCP) joints using a PolyU finger knuckle image contact-less database.
\subsection{Feature Extraction and Classification}
Earlier studies in the literature explored to identify the finger knuckle patterns formed between the middle phalanx and proximal phalanx bones. The first study named as competitive code focused to use major knuckle near PIP joint employed 2D Gabor filter to extract orientation information \cite{20}. In \cite{22}, authors resolved the problems occurring in finger knuckle recognition due to pose variations or the presence of artifacts. In \cite{21}, authors made efforts to improve the knuckle code approach by applying radon transform on enhanced knuckle images and achieved 1.08 \% EER and 98.6 \% rank one recognition rate. In \cite{49}, authors fused hand vein and knuckle shape features to authenticate the individuals. In \cite{25}, authors incorporated the phase and orientation of knuckle features. In \cite{36}, the first study that claimed about the significance of minor finger knuckle patterns near DIP joint for human identification. In \cite {38}, the authors considered recovered minutiae of knuckle samples for template matching using a data-set of 120 subjects. In \cite{6}, the authors made an attempt to highlight minor lower finger knuckle patterns (MCP joint) between the proximal and the metacarpal phalanx bones of fingers. In \cite{58}, efforts were made to explore the overall information present over hand dorsal surface. The lower minor knuckle and palm dorsal texture were fused using two separate data-sets and the results signified the potential of dorsal texture as a biometric identifier. In another study \cite{69}, the authors presented a score level fusion of multiple texture features obtained by local transformations schemes and then applied non-rigid matching criterion. In a recent study \cite{70}, finger dorsal cancelable templates were generated to show the uniqueness and stability of finger knuckle patterns in real world scenarios.  


\section{Proposed Siamese based CNN  Network}
In this paper, we propose a siamese based CNN matching framework, FKIMNet.  First, we have extracted ROI's of the major knuckle, minor knuckle and nail traits over PolyU FKI dataset by training a state-of-the-art region based convolution neural network (CNN) that uses different bounding boxes as ground truth to classify and localize the ROI's \cite{63}. For any given image $I$, the network returns a 128-D feature embedding $\theta (I)$. For matching any 2 images $I_p$ and $I_q$, we compute the corresponding feature embedding $\theta (I_p)$ and $\theta (I_q)$ and calculate the $L_2$ distance between them. Where, $d = L_2(\theta (I_p), \theta (I_q))$, is the similarity score between the images $I_p$ and $I_q$. The value of $d$ should be closer to $0$ for images belonging to the same subject whereas, it should be high for images of different subjects. To ensure such behaviour, we train the network using triplet loss function. For efficient and effective training of the network, we nourish it by incorporating dynamically adaptive margin, hard negative mining and data augmentation.

\subsection{Model Architecture}

\begin{figure*}[!htp]
\small
\centering
\includegraphics[width=1\linewidth]{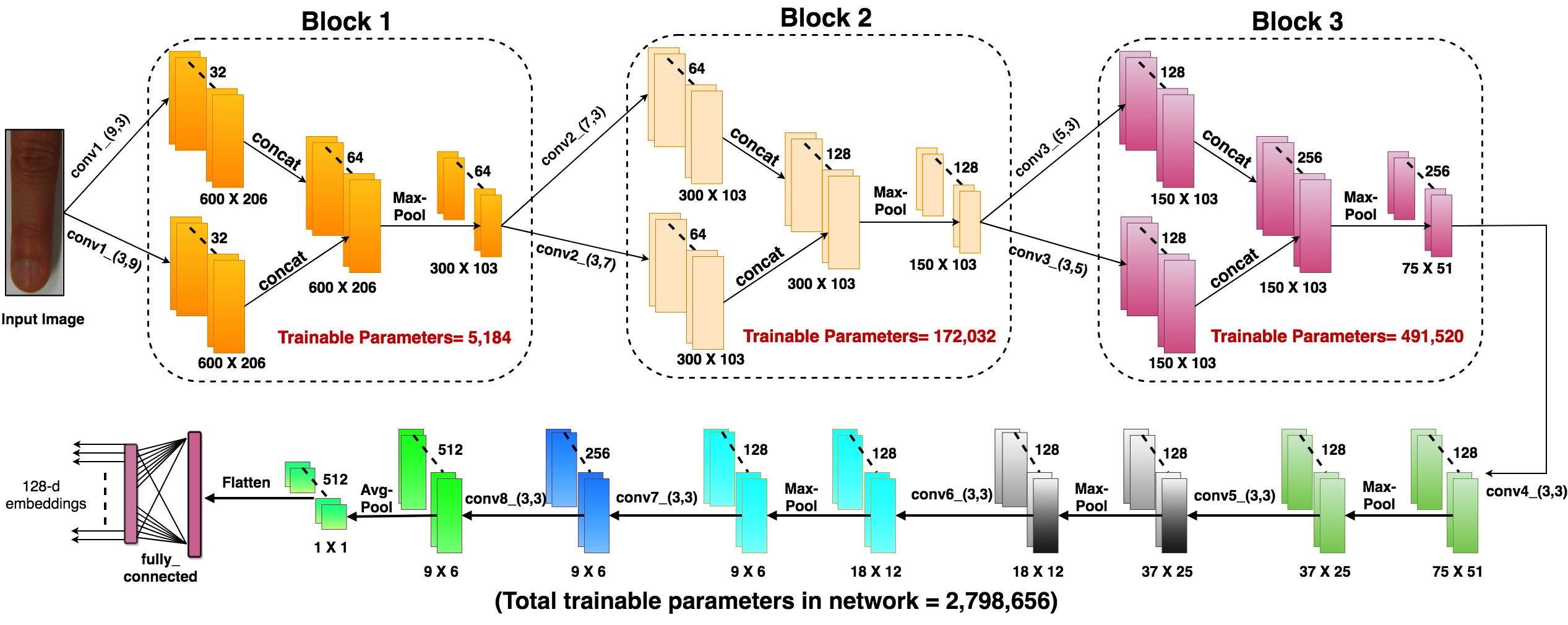} %
\caption{Proposed FKIMNet N/W Architecture: Feature Extraction}
\label{fig:ac_1}
\end{figure*}

Contrary to popular deep learning works which use pre-trained deep architectures like VGG \cite{66}, Inception \cite{67} or ResNet \cite{68} for feature extraction, we propose a novel convolutional neural network. Since deep networks require a large amount of data for tuning the parameters, they readily over-fit when data is less. That is the training loss becomes very low whereas the testing loss is still high. As the training loss becomes negligible, the network is not able to learn any further and does not perform well in testing data. As the number of images in the knuckle dataset is low (only 2 images per subject), we devised a small siamese model to avoid over-fitting. The network details are shown in Fig. \ref{fig:ac_1}. 

The major challenge in matching knuckle images is non-rigid distortion. Due to its presence, small convolutional filters in initial layers like $3\times3$, tend to give varying activation maps, resulting in an immense difference in feature embedding of same class images. Hence, we have applied large filters, achieving invariance to such local distortions. Most of the features in Knuckle images are line-based features. To capture them, the network has horizontal and vertical filters. Horizontal filters are targeted to capture the lines in the image. Vertical filters detect the spatial relationship between different lines. On a given input, we apply both horizontal and vertical filters and concatenate the output produced by both of them. Over this, we apply max pooling to decrease the feature size. 

In Fig.\ref{fig:ac_1} it can be seen that the network has $3$ blocks of horizontal and vertical filters with max pooling in the initial layers. The feature map produced after the $3^{rd}$ block now contains aggregated global information in it. That is all the local features such as lines and distortions are no longer prominent. Hence, all the later convolution layers of the model have standard $3\times3$ filters in them. As the finger images are highly rectangular in size, the $2^{nd}$ block of the network only pools the feature map in the horizontal direction. Finally, the network concludes with a fully connected layer of $128$ neurons, giving us a $128$-D feature embedding for each image. The embedding are normalised as to lie on a zero mean unit norm hypersphere. The network in total only has $2.8$ million trainable parameters which allow the network to be generalized across testing data as well.

\begin{figure}[htb!]
\small
\centering
\includegraphics[width=1\linewidth]{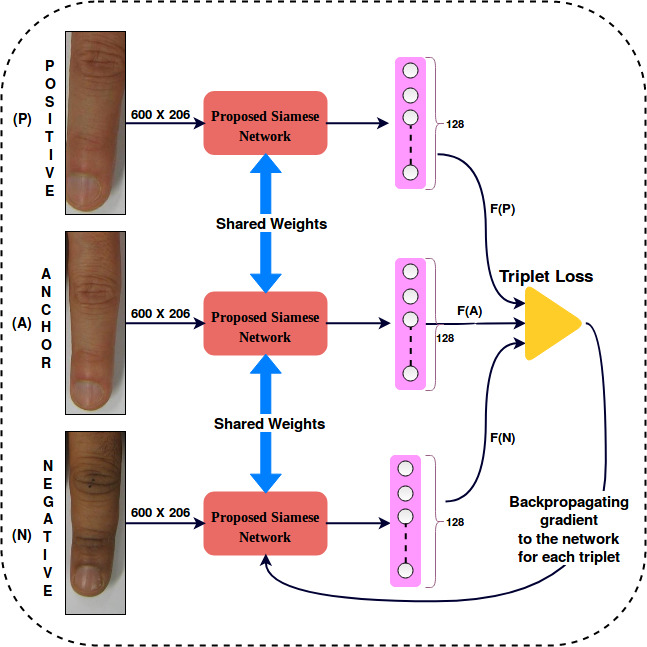} %
\caption{Proposed FKIMNet N/W: Triplet Loss}
\label{fig:ac}
\end{figure}

\subsection{Network Training}
For training aforementioned model, we use triplet loss function as described below. We have found out that training such networks is not a trivial task, hence efficient and effective batch making plays a vital role in training. To make the batching robust we have used data augmentation and hard negative mining, ensuring the proper convergence of the network. For better accuracy of the model, we have also utilized a dynamic adaptive margin.

\subsubsection{Robust Training via. Triplet Loss}
For a given image $I$, the network produces an embedding $\theta (I)$, such that given two images $I_p$ and $I_q$, if $I_p$ and $I_q$ belong to the same subject, then $L_2(\theta (I_p),\theta (I_q)) =0$, otherwise, $L_2(\theta (I_p),\theta (I_q)) \geq\beta$, where $\beta$ is the margin. For making the embedding of same class images similar, we can minimize the Euclidean loss between the embedding as shown as:
\begin{equation}
    L_p=\frac{1}{N}\sum_{i=1}^{N}[\underbrace{(\theta_i(I_p)-\theta_i(I_q))^2}_\text{$L_2$ Distance between embeddings}]
\end{equation}
For making the embeddings of different class images as far as possible, we can minimize the hinge loss between the embeddings as shown as:
\begin{equation}
L_n=\frac{1}{N}\sum_{i=1}^{N}[max(0,\underbrace{\beta-(\theta_i(I_p)-\theta_i(I_q))^2}_\text{Deviation of $L_2$ Distance from the margin})]
\end{equation}
Since both the tasks are needed to be accomplished simultaneously, we combine both the losses to form a single triplet loss. It is defined over $3$ embeddings, $\theta(I_a)$: embedding of anchor image, $\theta(I_p)$: embedding of positive image and $\theta(I_n)$: embedding of negative image as shown as:
{\begin{equation}
\begin{split}
    L = &\frac{1}{N}\sum_{i=1}^{N} \\&[max(0,  \underbrace{( \theta_{i}(I_a) - \theta_{i}(I_p) )^{2}}_\text{For anchor and positive images} - \underbrace{( \theta_{i}(I_a) - \theta_{i}(I_n))^{2}  + \beta}_\text{For anchor and negative images})]
\end{split}
\end{equation}}

\begin{figure}[htb!]
\small
\centering
\includegraphics[width=0.92\linewidth,height=0.75\linewidth]{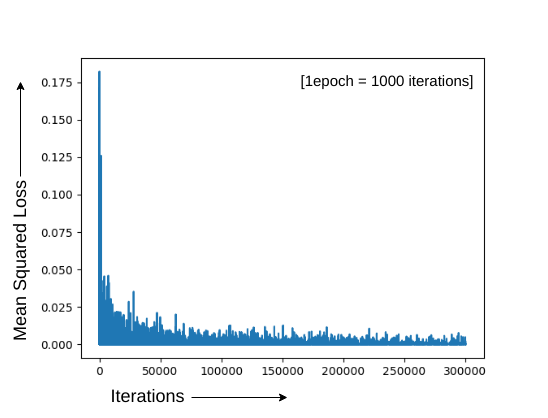} %
\caption{Training loss for FKIMNet over full finger data. Usually due to easy triplets Siamese loss drastically goes down and fails to generalize. Adaptive margin and hard triplet mining introduces multiple spikes through out the training iterations enabling the network to generalize.}
\label{fig:los}
\end{figure}

The overall training loss has been shown in Fig. \ref{fig:los}. We have observed that the loss of the network decreases immediately in the first few epochs and gradually converges in the subsequent epochs, depicting a healthy training regime for the network.

\subsubsection{Efficient and Effective Batch Making}
For training the network, we provide triplets as shown in Fig. \ref{fig:ac}. A stack of $3$ images is provided in which one of them is the anchor, other is positive and one is the negative image. Choosing these triplets is a hard task as there is a huge difference in the number of positive and negatives pairs (ratio of $1:502$). As the number of positive pairs is very low, we use data augmentation to increase their quantity. Since, negative pairs dominate the dataset, choosing relevant negative images is of utmost importance. To achieve that we use hard negative mining.

\begin{figure}[htb!]
\small
\centering
\includegraphics[width=0.82\linewidth,height=0.92\linewidth]{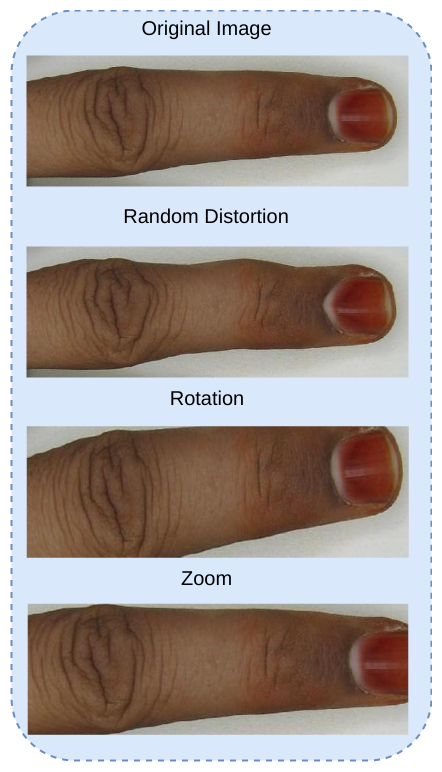} %
\caption{Data augmentation over full finger data enables to introduce more and more variety in training triplets.}
\label{fig:ac1}
\end{figure}

\textbf{Data Augmentation for Generalized Training}: The lack of variation in the training data, in terms of zoom, rotation etc could lead to a less robust and less scalable system. Hence, the 2 original training images per subject are augmented, to now have 35 training images by introducing random zoom, distortions and rotational variance into them \cite{2}. The need for these augmentations come from the challenges in the practical implementation of our model. The real-world images might be at a different angle, or have a different degree of zoom or have distortions due to various physical conditions. The augmentation, thus, incorporates these variations into the training leading to a better test performance and hence a more robust model. Fig. \ref{fig:ac1} shows augmented images on PolyU FKI dataset.

\textbf{Hard Negative Mining}: While forming a triplet for training, choosing a suitable negative pair is a demanding job. Given the enormous amount of negative pairs, choosing them randomly will lead to the formation of easy triplets. The network will easily learn these triplets ignoring the challenging ones, degrading the network's performance. To avoid that, we have to choose the hard triplets. A triplet is said to be hard when the distance between the embeddings of anchor and negative ($d_n$), and anchor and positive ($d_p$) is lesser than $\beta$. To compute such triplets, we have to calculate the embeddings of each image in the dataset before making every batch, which can be a cumbersome task. Hence, while creating a batch, we randomly choose $1000$ triplets, compute $d_n$ and $d_p$ for each triplet and only choose those whose $d_n-d_p\leq\beta$ for batch making.

Fig. \ref{fig:los} shows the plot of mean squared error with respect to the iterations when the network was trained on full finger images. Since each of the embeddings is projected on a unit norm hypersphere, the initial loss is approximately $0.175$ only. As the training starts, every triplet violates the margin constraint, hence the loss decreases rapidly. Now after few iterations, a majority of the triplets have become easy triplets, hence choosing triplets randomly will result in the saturation of the loss, causing the network to not learn any further. Whereas, with hard triplets, those examples are chosen which produce a larger loss. As shown in Fig. \ref{fig:los}, the frequent spikes in the loss are due to the hard samples. As these samples do not let the loss of the network to saturate, the network is trained properly and is highly generalized.

\begin{figure}[htb!]
\small
\centering
\includegraphics[width=0.92\linewidth]{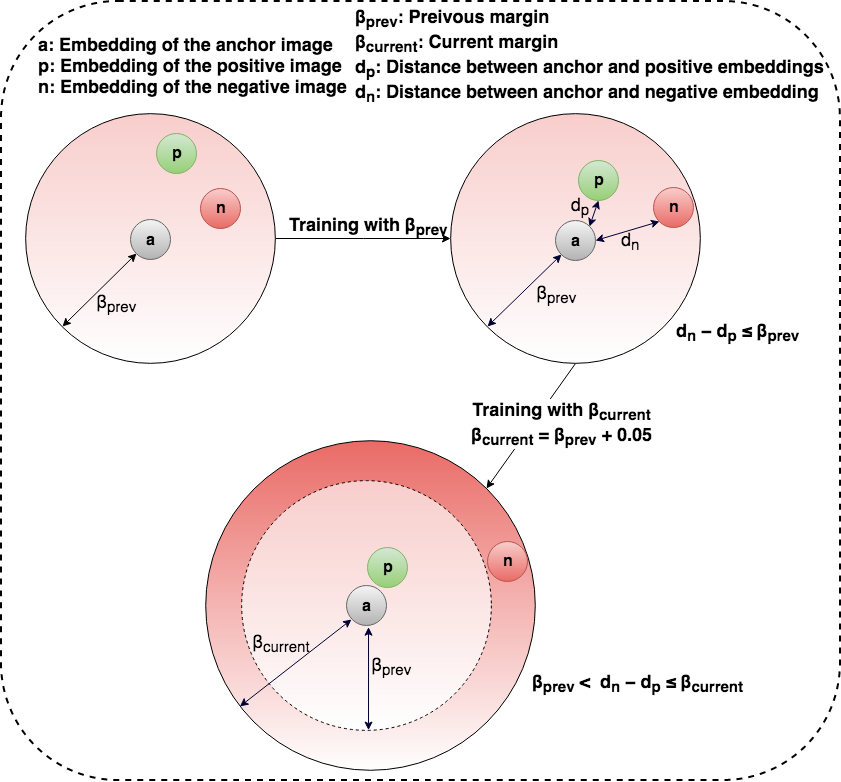} %
\caption{Dynamic adaptive margin used for discrimination and better network generalization and convergence.}
\label{fig:ac_am}
\end{figure}

\textbf{Adaptive Margin for Better Discrimination:} As the training process progresses, the number of hard triplets reduce. This behavior can be accounted to the selection process of triplets. As the model trains, the embeddings of the anchor, positive and negative differ with every epoch such that $d_p$ decreases and $d_n$ increases. A triplet which was considered hard in the initial epochs will not be considered the same in the later epochs as the distance between their embeddings will differ. In order to overcome this problem, we use the concept of adaptive margin. As the model trains, we increase our margin by a step. As shown in the Fig. \ref{fig:ac_am}, in the early training process, $d_n - d_p < \beta_{prev}$, where $\beta_{prev}$ is the previous margin. But as the model trains, $d_n - d_p$ crosses margin. With increasing the margin, we get $\beta_{current} >  d_n - d_p > \beta_{prev}$, where $\beta_{current}$ is the current margin such that $\beta_{current}=\beta_{prev}+0.05$.

\textbf{Hyper-Parametrization:} The model architectural parameters are shown in Fig. \ref{fig:ac_1}. The network has been trained using the Adam optimizer with its default parameters. The batch size used is $105$ and the models have been trained for $500$ epochs. Initially, the margin $\beta$ is set at $0.2$, while it dynamically increases by $0.05$ values to a maximum of $0.5$. ReLU activation function has been applied over each layer except the last. The output of the last layer is normalized onto a zero centred unit radius $128$-D hyper-sphere.


\section{Experimental Analysis}
\label{sec:EA}
In distinguishing experiments, the performance of the proposed approach has been evaluated in terms of EER (Equal Error Rate), CRR (Correct Recognition Rate) and DI (Decidability Index) for rigorous testing. It is to be noted that we have only considered Rank-1 accuracy (CRR). All these performance parameters are defined in \cite{39}.

\textbf{Database Specifications :} The two publicly available finger knuckle image databases i.e., PolyU FKP \cite{55}, and PolyU Contactless FKI \cite{56} have been used for the analysis and validation of the proposed approach. However, the images also suffer from finger artifacts, low contrast, illumination variation, reflection, and non-rigid deformations. The PolyU FKI consists of 5 finger images per subject for all the 503 classes. While in case of PolyU FKP dataset, the 24 left and 24 right finger knuckle samples are collected from each of 165 subjects, which resulted in 7920 images. It is to be noted that, the PolyU FKI dataset \cite{56} has a complete finger dorsal images over which major knuckle, minor knuckle and nail ROI's are extracted. On the other hand, PolyU FKP dataset \cite{55} only provides major finger knuckle images. 

\textbf{Testing Protocol :} Since both of the above-discussed databases have different images and class numbers. Therefore we have employed multiple testing protocols for rigorous testing. During testing on PolyU FKP \cite{55}, the first six samples per subject are considered for training and the remaining six for testing. Thus, we obtained $23,760$ genuine matching and $15,657,840$ impostor matching scores. On the other hand in case of PolyU FKI database, first two samples per subject are selected for training and rest three for testing. Therefore, we obtain $2,012$ genuine matching scores and $1,010,024$ impostor matching scores.

\begin{small}
\begin{table}[!htp]\scriptsize
\centering
\caption{ROC Performance Analysis for Unimodal Systems}
\label{tab:fktable1}
\begin{tabular}{|p {1.5 cm}|p {1.3cm}|p {1.6cm}|p {0.7cm}|p {0.7cm}|p {0.4cm}|}
\hline
\textbf{Traits} & \textbf{Dataset}& \textbf{Method}  & \textbf{EER} & \textbf{CRR} & \textbf{DI}  \\ \hline
\multicolumn{6}{|c|}{\textbf{Finger Components Matching (EER/CRR in \%)}: PolyU FKI Dataset} \\ \hline
Major Knuckle & Cropped \cite{63}   &   FKIMNet    &                 4.17      & 90.06        & 2.08   \\ \hline
Minor Knuckle & Cropped \cite{63}     &FKIMNet &   3.14  & 88.80          &  2.14   \\ \hline
Nail &  Cropped \cite{63}         & FKIMNet&   2.24     & 94.83  & 2.30    \\ \hline
Major Knuckle  & Standard \cite{56}& FKIMNet&     3.97    & 90.52  & 2.09   \\ \hline
Minor Knuckle & Standard   \cite{56}  & FKIMNet   &     3.36      &88.73    & 2.10    \\ \hline
\multicolumn{6}{|c|}{\textbf{Major FKP Matching}: PolyU FKP Dataset}                                      \\ \hline
Major FKP & Standard \cite{55}       &FKIMNet &2.03   & 94.02   &1.93\\ \hline
\multicolumn{6}{|c|}{\textbf{Finger Holistic Matching}: PolyU FKI Dataset}                                      \\ \hline
Full Finger& Standard \cite{56}    &  FKIMNet & 0.93    &97.01     & 2.41    \\ \hline
Full Finger& Standard \cite{56}   &  FaceNet \cite{64}  & 2.62   &80.78     & 2.15    \\ \hline
\multicolumn{6}{|c|}{\textbf{State-of-Art Performance Analysis}: PolyU FKI Dataset}                                      \\ \hline
 Major Knuckle   & Standard \cite{56} &  Log-Gabor \cite{6}    & 9.41 & NA       &  1.82\\ \hline
 Minor Knuckle & Standard \cite{56} &  Log-Gabor  \cite{6}  & 12.60 &   NA  &  1.65\\ \hline
Major-Minor &Standard \cite{56} &  Log-Gabor \cite{6}&     5.92 & NA          &  2.85\\ \hline
\multicolumn{6}{|c|}{\textbf{State-of-Art Performance Analysis}: PolyU FKP Dataset}  \\ \hline
  Major Knuckle &Standard \cite{55} &  MexCode  \cite{34}    & 1.82 & NA     &  2.02\\ \hline
  Major Knuckle &Standard \cite{55} & MrMxCode \cite{34}&    1.048 & NA        &  2.02\\ \hline
\end{tabular}
\end{table}
\end{small}

\begin{small}
\begin{table}[!htp]\scriptsize
\centering
\caption{ROC Performance Analysis for Multimodal Fusion}
\label{tab:fusion1}
\begin{tabular}{|l|l|l|l|l|}
\hline
\textbf{Traits} & \textbf{Method} & \textbf{EER} (\%)  & \textbf{CRR} (\%)  & \textbf{DI}  \\ \hline

\multicolumn{5}{|c|}{\textbf{PolyU FKI Database \cite{56}} }                                     \\ \hline
Major-Minor     &  FKIMNet & 2.22   & 94.30  & 2.38  \\ 
\hline
Major-Nail     &FKIMNet  & 1.35   & 97.21  & 2.58   \\ 
\hline
Minor-Nail     & FKIMNet  & 1.19   & 97.15  & 2.63   \\ 
\hline
Major-Finger     & FKIMNet  & 0.927   & 97.94  & 2.55   \\ 
\hline
Minor-Finger     & FKIMNet  & 0.722   & 97.68  & 2.50   \\ 
\hline
Nail-Finger      & FKIMNet & 0.728   & 98.40  & 2.63   \\ 
\hline
Major-Minor-Nail    &  FKIMNet   & 0.696   & 98.277  & 2.76    \\ 
\hline
Major-Minor-Finger   &  FKIMNet  & 0.76 & 98.07          &  2.65 \\
\hline
Major-Nail-Finger   & FKIMNet & 0.62 & 98.74          &  2.78 \\
\hline
Minor-Nail-Finger   & FKIMNet & 0.56& 98.86          &  2.80 \\
\hline
Maj-Min-Nail-Finger   &  FKIMNet  & 0.40 & 98.60          &  2.85\\
\hline
\end{tabular}
\end{table}
\end{small}

\subsection{Finger Components based Matching}
\begin{figure*}[!hbtp]
\small
\centering
\subfloat[Single modality based performance]{\includegraphics[scale = 0.47]{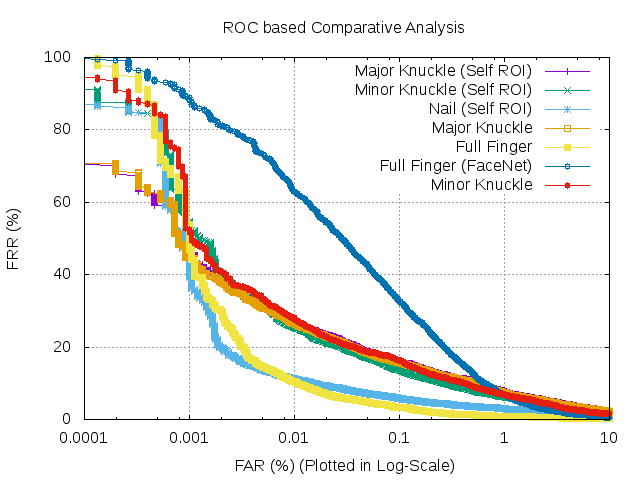}} %
\hspace{1 em}
\subfloat[Fusion based performance]{\includegraphics[scale = 0.47]{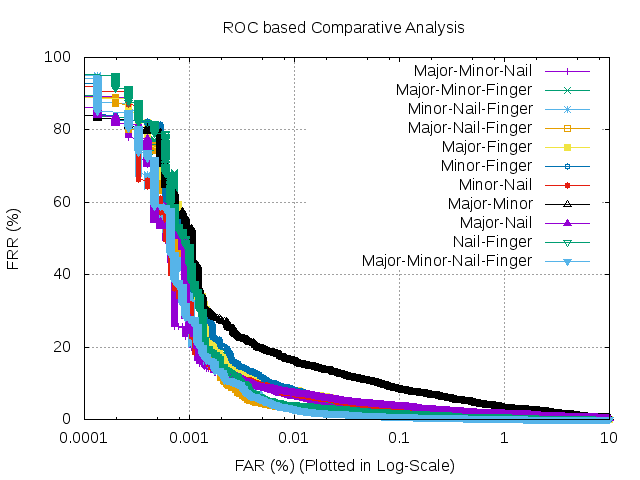}} %
\caption{ROC based Performance Analysis}
\label{fig:uni1}
\end{figure*}

We carry out extensive ablation experiments to analyze the potential of proposed FKIMNet on each finger component. The corresponding ROC characteristics w.r.t Polyu FKP and PolyU FKI datasets are shown in Figure \ref{fig:uni1}(a). Some promising conclusions are summed up next according to the ablative results listed in Table \ref{tab:fktable1}. (1) The comparison between the first three rows indicates that fingernail is the best performing finger component than major and minor knuckle. In particular, we achieve best performance results on the nail with EER of 2.24 \% and CRR of 94.83 \%. This is because ROI segmentation of the nail is very consistent due to its unique shape and FKIMNet performs very well over it. (2) The proposed adaptive triplet loss function effectively improves the performance of given state-of-art major and minor knuckle ROI's as indicated in fourth and fifth rows of Table \ref{tab:fktable1}. This is because the given ROI's of major and minor knuckles \cite{56} are consistent than our automated extracted ROI's. Secondly, for this case, the proposed model is trained and tested on same standard images \cite{56}, while in the previous case, the model is trained on standard images \cite{56} but tested on self-extracted ROI images. (3) In addition, we illustrate the effectiveness of proposed FKIMNet on one another publicly available dataset \cite{55}. A lower value of EER i.e., 2.03 \% has been achieved on major FKP, which is superior to the EER values obtained from any of the above-mentioned cases.

\subsection{Full Finger Matching}
In the second experiment, we consider the full dorsal finger as given in \cite{56} for testing. This is to check the robustness of proposed FKIMNet as well as to justify our assumption that, finger contains more discriminative dorsal textures than major and minor knuckle patterns. The different testing protocols have been listed in Table \ref{tab:fktable1}. It is to be noted that the proposed approach has surpassed all the previous cases when tested on full finger dorsal image using the same testing protocol that employs 2 training and 3 testing images per subject. One can see higher recognition results i.e., EER of 0.928 \% and CRR of 97.01 \%, when FKIMNet is trained and tested with all 503 images. In comparison to this, another state-of-art deep network architecture named as FaceNet \cite{64} has been tested using same testing protocols, tweaked and fine tuned as required. In additional experimentation, when we trained our model on only 250 subjects (zero shot learning) and tested it on 503 subjects, a small drop (EER of 4.438\%) in results can be seen. But this performance deviation is still better than the similar works done in literature. Similarly, a testing strategy harder than previous once is devised to make a more fair comparison which employs 1 training and 4 testing images. By doing so, the performance (EER of 8.007\%) falls abruptly and found lower than all the studied cases. This justifies the strength of FKIMNet as well as the importance of full finger texture. 


\subsection{Fusion and Comparative Analysis}
In this test, we present fusion of various finger traits at a weighted sum score level fusion. The respective ROC curves for various fusion combinations are shown in Figure \ref{fig:uni1}(b). In addition, we compare the performance with well known state-of-art methods. Table \ref{tab:fusion1} illustrates the comparative performance analysis as discussed in Section \ref{sec:EA}. It can be seen that the proposed multimodal fusion of two, three and four traits surpass other single modality combinations as well as state-of-art methods \cite{6, 34}. We also found that CRR increases in all studied cases but not much at Rank-1 because FKIMNet requires right positive-negative image pairs to learn distance margin. On the contrary, there are certain ROI's which are not properly segmented, thus our model doesn't effectively match the underlying features at Rank-1.
\section{Conclusion} 
This work demonstrates the feasibility of full finger dorsal texture (holistic matching) for personal recognition system. In this work, we have introduced nail texture/shape for the first time. We have shown that, it is the best performing finger component better than major and minor, primarily due to better nail registration/alignment. The proposed FKIMNet shows quite impressive results on publicly available finger knuckle image databases namely PolyU FKP and PolyU Contact-less FKI. We have also demonstrated that holistic finger image matching is better than any other finger component matching again due to better ROI alignment. We have performed rigorous experiments on three finger dorsal components (major knuckle, minor knuckle and nail) along with full finger and their various fusions. Such basic component/full finger based score level fusion in pursuit of better performance produces very good results and nicely augments the justification of the proposed FKIMNet and the suitability of finger knuckle dorsal images. 


{\small
\bibliographystyle{IEEEtran}
\bibliography{gaurav}
}
\end{document}